\documentclass[sigconf]{acmart}

\copyrightyear{2023}
\acmYear{2023}
\setcopyright{rightsretained}
\acmConference[KDD '23]{Proceedings of the 29th ACM SIGKDD Conference on Knowledge Discovery and Data Mining}{August 6--10, 2023}{Long Beach, CA, USA}
\acmBooktitle{Proceedings of the 29th ACM SIGKDD Conference on Knowledge Discovery and Data Mining (KDD '23), August 6--10, 2023, Long Beach, CA, USA}
\acmDOI{10.1145/3580305.3599347}
\acmISBN{979-8-4007-0103-0/23/08}

\AtBeginDocument{%
	\providecommand\BibTeX{{%
			\normalfont B\kern-0.5em{\scshape i\kern-0.25em b}\kern-0.8em\TeX}}}
\usepackage{caption}
\usepackage{subcaption}
\usepackage{appendix}
\usepackage{flushend}
\usepackage{balance}
\usepackage{lineno}
\usepackage{amsmath,amsfonts}
\usepackage{algorithm}
\usepackage{algorithmic}

\usepackage{graphicx}
\usepackage{textcomp}
\usepackage{xcolor}
\usepackage{amsthm}
\usepackage{url}
\usepackage{float}
\usepackage{multirow}
\usepackage{multicol}
\usepackage{color}
\usepackage{bm}
\usepackage{bbm}

\newtheorem{definition}{Definition}

\usepackage{pifont}

\newcommand{\bp}{\mathbf{p}}
\newcommand{\bq}{\mathbf{q}}

\newcommand{\bH}{\mathbf{H}}
\newcommand{\bh}{\mathbf{h}}

\newcommand{\mcT}{\mathcal{T}}
\newcommand{\mcC}{\mathcal{C}}

\newcommand{\bz}{\mathbf{z}}

\usepackage{array}
\newcolumntype{L}[1]{>{\raggedright\let\newline\\\arraybackslash\hspace{0pt}}m{#1}}
\newcolumntype{C}[1]{>{\centering\let\newline  \\\arraybackslash\hspace{0pt}}m{#1}}
\newcolumntype{R}[1]{>{\raggedleft\let\newline \\\arraybackslash\hspace{0pt}}m{#1}}

\DeclareMathOperator*{\argmax}{argmax}
\DeclareMathOperator*{\argmin}{argmin}

\copyrightyear{2023}
\acmYear{2023}
\setcopyright{rightsretained}
\acmConference[KDD '23]{Proceedings of the 29th ACM SIGKDD Conference on Knowledge Discovery and Data Mining}{August 6--10, 2023}{Long Beach, CA, USA}
\acmBooktitle{Proceedings of the 29th ACM SIGKDD Conference on Knowledge Discovery and Data Mining (KDD '23), August 6--10, 2023, Long Beach, CA, USA}
\acmDOI{10.1145/3580305.3599347}
\acmISBN{979-8-4007-0103-0/23/08}


\begin{document}
\title{Federated Few-shot Learning}

	\author{Song Wang}
\affiliation{%
  \institution{University of Virginia} \country{}}
\email{sw3wv@virginia.edu}
	
	\author{Xingbo Fu}
\affiliation{%
  \institution{University of Virginia} \country{}}
\email{xf3av@virginia.edu}

		\author{Kaize Ding}
\affiliation{%
  \institution{Arizona State University}  \country{}}

\email{kaize.ding@asu.edu}

				\author{Chen Chen}
\affiliation{%
  \institution{University of Virginia} \country{}}
\email{zrh6du@virginia.edu}

				\author{Huiyuan Chen}
\affiliation{%
  \institution{Case Western Reserve University} \country{}}
\email{hxc501@case.edu}	

					\author{Jundong Li}
\affiliation{%
  \institution{University of Virginia} \country{}}
\email{jundong@virginia.edu}

\renewcommand{\shortauthors}{Song Wang et al.}
\begin{abstract}
    Federated Learning (FL) enables multiple clients to collaboratively learn a machine learning model without exchanging their own local data. In this way, the server can exploit the computational power of all clients and train the model on a larger set of data samples among all clients. 
    Although such a mechanism is proven to be effective in various fields,
    existing works generally assume that each client preserves sufficient data for training. In practice, however, certain clients may only contain a limited number of samples (i.e., few-shot samples). For example, the available photo data taken by a specific user with a new mobile device is relatively rare. In this scenario, existing FL efforts typically encounter a significant performance drop on these clients. 
    Therefore, it is urgent to develop a few-shot model that can generalize to clients with limited data under the FL scenario. In this paper, we refer to this novel problem as \emph{federated few-shot learning}.
    Nevertheless, the problem remains challenging
    due to two major reasons: the global data variance among clients (i.e., the difference in data distributions among clients) and the local data insufficiency in each client (i.e., the lack of adequate local data for training). 
    To overcome these two challenges,
    we propose a novel federated few-shot learning framework with two separately updated models and dedicated training strategies to reduce the adverse impact of global data variance and local data insufficiency.
    Extensive experiments on four prevalent datasets that cover news articles and images validate the effectiveness of our framework compared with the state-of-the-art baselines. Our code is provided\footnote{\href{https://github.com/SongW-SW/F2L}{https://github.com/SongW-SW/F2L}}.
\end{abstract}

\maketitle

\section{Introduction}
The volume of valuable data is growing massively with the rapid development of mobile devices~\cite{liu2011web,chakrabarti2002mining}. 
Recently, researchers have developed various machine learning methods~\cite{chau2008machine,zhang2008web,xu2010web} to analyze and extract useful information from such large-scale real-world data. 
Among these methods, Federated Learning (FL) is an effective solution, which aims to collaboratively optimize a centralized model over data distributed across a large number of clients~\cite{collins2021exploiting,zhao2018federated,fallah2020personalized,kairouz2021advances}. In particular, FL trains a global model on a server by aggregating the local models learned on each client~\cite{briggs2020federated}. 
 Moreover, by avoiding the direct exchange of private data, FL can provide effective protection of local data privacy for clients~\cite{li2021survey}.
As an example, in Google Photo Categorization~\cite{duan2020self,lim2020federated}, the server aims to learn an image classification model from photos distributed among a large number of clients, i.e., mobile devices.
 In this case, FL can effectively conduct learning tasks without revealing private photos to the server. 
    
\begin{figure}[t]
					
	    \centering
	    \includegraphics[width=0.97\linewidth]{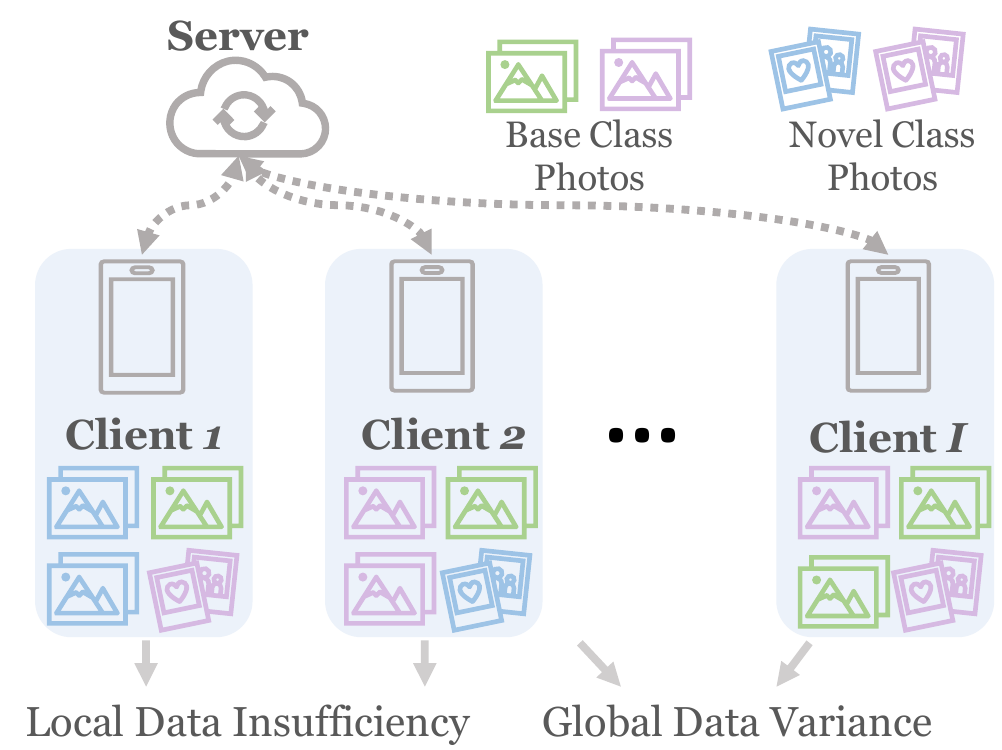}
	    \vspace{-0.5mm}
	    \caption{The two challenges of federated few-shot learning as an example in Google Photo Categorization: local data insufficiency and global data variance.}
	    \vspace{-3.5mm}
	    \label{fig:intro}
	  
	\end{figure}

In fact, new learning tasks (e.g., novel photo classes) are constantly emerging over time~\cite{tian2020rethinking,yao2021meta}. In consequence, FL can easily encounter a situation where the server needs to solve a new task with limited available data as the reference.
In the previous example of Google Photo Categorization, as illustrated in Fig.~\ref{fig:intro}, the server may inevitably need to deal with novel photo classes such as the latest electronic products, where only limited annotations are available. 
Nevertheless, existing FL works generally assume sufficient labeled samples for model training, which inevitably leads to unsatisfying classification performance for new tasks with limited labeled samples~\cite{fan2021federated}. 
Therefore, to improve the practicality of FL in realistic scenarios, it is important to solve this problem by learning an FL model that can achieve satisfactory performance on 
new tasks with limited samples. 
In this paper, we refer to this novel problem setting as \emph{federated few-shot learning}.

 
 

Recently, many few-shot learning frameworks~\cite{vinyals2016matching,finn2017model,snell2017prototypical,wang2022faith} have been proposed to deal with new tasks with limited samples. Typically, the main idea is to learn meta-knowledge from \emph{base classes} with abundant samples (e.g., photo classes such as portraits). Then such meta-knowledge is generalized to \emph{novel classes} with limited samples (e.g., photo classes such as new electronic products), where novel classes are typically disjoint from base classes. 
However, as illustrated in Fig.~\ref{fig:intro}, it remains challenging to conduct few-shot learning under the federated setting due to the following reasons. 
First,
due to the \emph{global data variance}
(i.e., the differences in data distributions across clients), the aggregation of local models on the server side will disrupt the learning of meta-knowledge in each client~\cite{karimireddy2020scaffold}. Generally, the meta-knowledge is locally learned from different classes in each client and thus is distinct among clients, especially under the non-IID scenario, where the data variance can be even larger among clients compared with the IID scenario. Since the server will aggregate the local models from different clients and then send back the aggregated model, the learning of meta-knowledge in each client will be potentially disrupted.
Second,
due to the \emph{local data insufficiency} in clients, it is non-trivial to learn meta-knowledge from each client. In FL, each client only preserves a relatively small portion of the total data~\cite{arivazhagan2019federated,fallah2020personalized}. However, meta-knowledge is generally learned from
data in a variety of classes~\cite{finn2017model,vinyals2016matching}. As a result, it is difficult to learn meta-knowledge from data with less variety, especially in the non-IID scenario, where each client only has a limited amount of classes.

To effectively solve the aforementioned challenges, we propose a novel \underline{\textbf{F}}ederated \underline{\textbf{F}}ew-shot \underline{\textbf{L}}earning framework, named F$^2$L. 
First, we propose a decoupled meta-learning framework to mitigate the disruption from the aggregated model on the server. Specifically, the proposed framework retains a unique \emph{client-model} for each client to learn meta-knowledge and a shared \emph{server-model} to learn client-invariant knowledge (e.g., the representations of samples), as illustrated in Fig.~\ref{fig:decouple}. 
Specifically, the client-model in each client is updated locally and will not be shared across clients, while the server-model can be updated across clients and sent to the server for aggregation. Such a design decouples the learning of meta-knowledge (via client-model) from learning client-invariant knowledge (via server-model). In this way, we can mitigate the disruption from the aggregated model on the server caused by global data variance among clients.
Second, to compensate for local data insufficiency in each client, we propose to leverage global knowledge learned from all clients with two dedicated update strategies.
In particular, we first transfer the learned meta-knowledge in client-model to server-model by maximizing the mutual information between their output (i.e., \textit{local-to-global knowledge transfer}). Then we propose a partial knowledge distillation strategy for each client to selectively extract useful knowledge from server-model (i.e., \textit{global-to-local knowledge distillation}). In this manner, each client can leverage the beneficial knowledge in other clients to learn meta-knowledge from more data.
In summary, our contributions are as follows:
\begin{itemize}
    \item \textbf{Problem}. We investigate the challenges 
    of learning meta-knowledge 
    in the novel problem of federated few-shot learning from the perspectives of \emph{global data variance} and \emph{local data insufficiency}. We also discuss the necessity of tackling these challenges.
    \item \textbf{Method}. We develop a novel federated few-shot learning framework F$^2$L with three essential strategies: (1) a decoupled meta-learning framework to mitigate disruption from the aggregated model on the server; (2) mutual information maximization for local-to-global knowledge transfer; (3) a novel partial knowledge distillation strategy for global-to-local knowledge distillation.
  \item \textbf{Experiments}. We conduct experiments on four few-shot classification datasets covering both news articles and images under the federated scenario. The results further demonstrate the superiority of our proposed framework.
\end{itemize}

\vspace{-0.1in}
\section{Preliminaries}
\subsection{Problem Definition}
In FL, given a set of $I$ clients, i.e., $\{\mathbb{C}^{(i)}\}_{i=1}^I$, where $I$ is the number of clients, each $\mathbb{C}^{(i)}$ owns a local dataset $\mathcal{D}^{(i)}$. The main objective of FL is to learn a global model over data across all clients (i.e., $\{\mathcal{D}^{(i)}\}_{i=1}^I$) without the direct exchange of data among clients.
Following the conventional FL strategy~\cite{mcmahan2017communication,fallah2020personalized,li2020federated}, a server $\mathbb{S}$ will aggregate locally learned models 
from all clients for a global model. 

Under the prevalent few-shot learning scenario,
we consider a supervised setting in which the data samples for client $\mathbb{C}^{(i)}$ are from its local dataset: $(x,y)\in\mathcal{D}^{(i)}$, where $x$ is a data sample, and $y$ is the corresponding label. We first denote the entire set of classes on all clients as $\mathcal{C}$. Depending on the number of labeled samples in each class, $\mathcal{C}$ can be divided into two categories: base classes $\mathcal{C}_b$ and novel classes $\mathcal{C}_n$,  where $\mathcal{C}=\mathcal{C}_b \cup \mathcal{C}_n$ and $\mathcal{C}_b \cap \mathcal{C}_n=\emptyset$. In general, the number of labeled samples in $\mathcal{C}_b$ is sufficient, while it is generally small in $\mathcal{C}_n$~\cite{finn2017model,snell2017prototypical}. Correspondingly, each local dataset can be divided into a base dataset $\mathcal{D}^{(i)}_b=\{(x,y)\in\mathcal{D}^{(i)}:y\in\mathcal{C}_b\}$ and a novel dataset $\mathcal{D}^{(i)}_n=\{(x,y)\in\mathcal{D}^{(i)}:y\in\mathcal{C}_n\}$.
In the few-shot setting, the evaluation of the model generalizability to novel classes $\mathcal{C}_n$ is conducted on $\mathcal{D}^{(i)}_n$, which contains only limited labeled samples. The data samples in $\mathcal{D}^{(i)}_b$ will be used for training. Then we can formulate the studied problem of federated few-shot learning as follows:

	\begin{definition}
	\textbf{Federated Few-shot Learning:} Given a set of $I$ clients $\{\mathbb{C}^{(i)}\}_{i=1}^I$ and a server $\mathbb{S}$,
federated few-shot learning aims to learn a global model after aggregating model parameters 
locally learned from $\mathcal{D}^{(i)}_b$ in each client such that the model can accurately predict labels for unlabeled samples (i.e., query set $\mathcal{Q}$) in $\mathcal{D}^{(i)}_n$ with only a limited number of labeled samples (i.e., support set $\mathcal{S}$).
	\end{definition}
	
	More specifically, 
 if the support set $\mathcal{S}$ consists of exactly $K$ labeled samples for each of $N$ classes from $\mathcal{D}^{(i)}_n$, and the query set $\mathcal{Q}$ is sampled from the same $N$ classes, the problem is defined as Federated $N$-way $K$-shot Learning. Essentially, the objective of federated few-shot learning is to learn a globally shared model across clients that can be fast adapted to data samples in $\mathcal{D}^{(i)}_n$ with only limited labeled samples. Therefore, the crucial part is to effectively learn meta-knowledge from the base datasets $\{\mathcal{D}_b^{(i)}\}_{i=1}^I$ in all clients. Such meta-knowledge is generalizable to novel classes unseen during training and thus can be utilized to classify data samples in each $\mathcal{D}^{(i)}_n$, which consists of only limited labeled samples.
	
\subsection{Episodic Learning}
\label{sec:episodic}
	
	In practice, we adopt the prevalent episodic learning framework for model training and evaluation, which has proven to be effective in various few-shot learning scenarios~\cite{vinyals2016matching,oreshkin2018tadam,ding2020graph,wang2021reform,ding2021few}. Specifically, the model evaluation (i.e., meta-test) is conducted on a certain number of \emph{meta-test tasks}, where each task contains a small number of labeled samples as references and unlabeled samples for classification. The local model training (i.e., meta-training) process in each client is similarly conducted on a specific number of \emph{meta-training} tasks, where each task mimics the structure of meta-test tasks. 
It is worth mentioning that meta-training tasks are sampled from the local base dataset $\mathcal{D}^{(i)}_b$ of each client, while meta-test tasks are sampled from the local novel dataset $\mathcal{D}^{(i)}_n$. That being said, the class set of samples in meta-training tasks is a subset of $\mathcal{C}_b$, while the class set of samples in meta-test tasks is a subset of $\mathcal{C}_n$, which is distinct from $\mathcal{C}_b$. The main idea of federated few-shot learning is to preserve the consistency between meta-training and meta-test so that the model can learn meta-knowledge from clients for better generalization performance to novel classes $\mathcal{C}_n$. 
	 
Specifically, to construct a meta-training task $\mathcal{T}$
in client $\mathbb{C}^{(i)}$ from its local base dataset $\mathcal{D}^{(i)}_b$, 
we first randomly sample $N$ classes from $\mathcal{D}^{(i)}_b$. Then we randomly select $K$ samples from each of the $N$ classes (i.e., $N$-way $K$-shot) to establish the support set $\mathcal{S}$. Similarly, the query set $\mathcal{Q}$ consists of $Q$ different samples (distinct from $\mathcal{S}$) from the same $N$ classes. The components of the meta-training task $\mathcal{T}$ is formulated as follows: 
\begin{equation}
    \begin{aligned}
    \mathcal{S}&=\{(x_1,y_1),(x_2,y_2),\dotsc,(x_{N\times K},y_{N\times K})\},\\
    \mathcal{Q}&=\{(q_1,y'_1),(q_2,y'_2),\dotsc,(q_{Q},y'_{Q})\},\\
    \mathcal{T}&=\{\mathcal{S},\mathcal{Q}\},
    \end{aligned}
\end{equation}
where $x_i$ (or $q_i)$ is a data sample in the sampled $N$ classes, and $y_i$ (or $y'_i$) is the corresponding label.
Note that during meta-test, 
each meta-test task shares a similar structure to meta-training tasks, except that the samples are from the local novel dataset $\mathcal{D}^{(i)}_n$, which are distinct from $\mathcal{D}^{(i)}_b$.

	\section{Methodology}

\begin{figure}[t]
					
	    \centering
	    \includegraphics[width=0.92\linewidth]{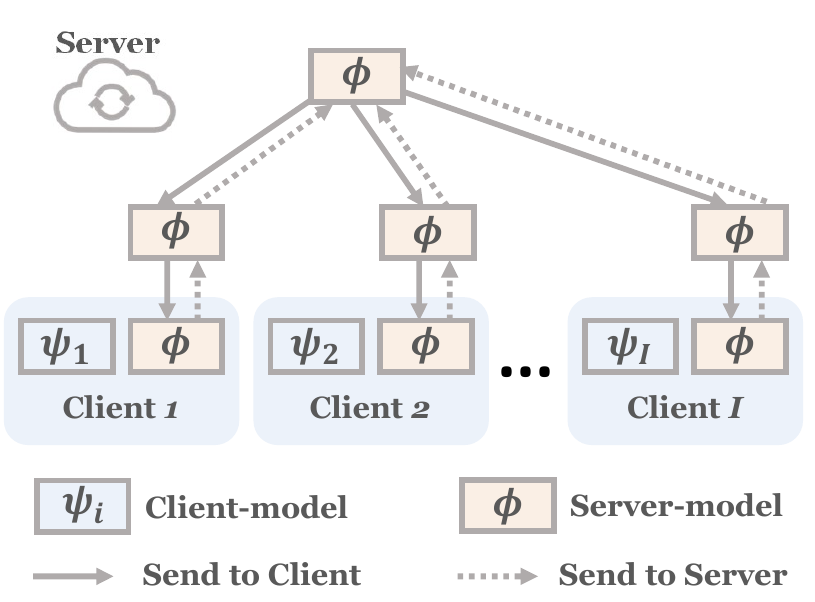}
	    \vspace{-1.5mm}
	    \caption{The illustration of our decoupled meta-learning framework. $\psi$ denotes the client-model, which will be locally kept in each client. $\phi$ denotes the server-model, which will be aggregated and sent to the server.}

	    \label{fig:decouple}
	  
	\end{figure}

\begin{figure*}[htbp]
	    \centering
	    \includegraphics[width=0.95\textwidth]{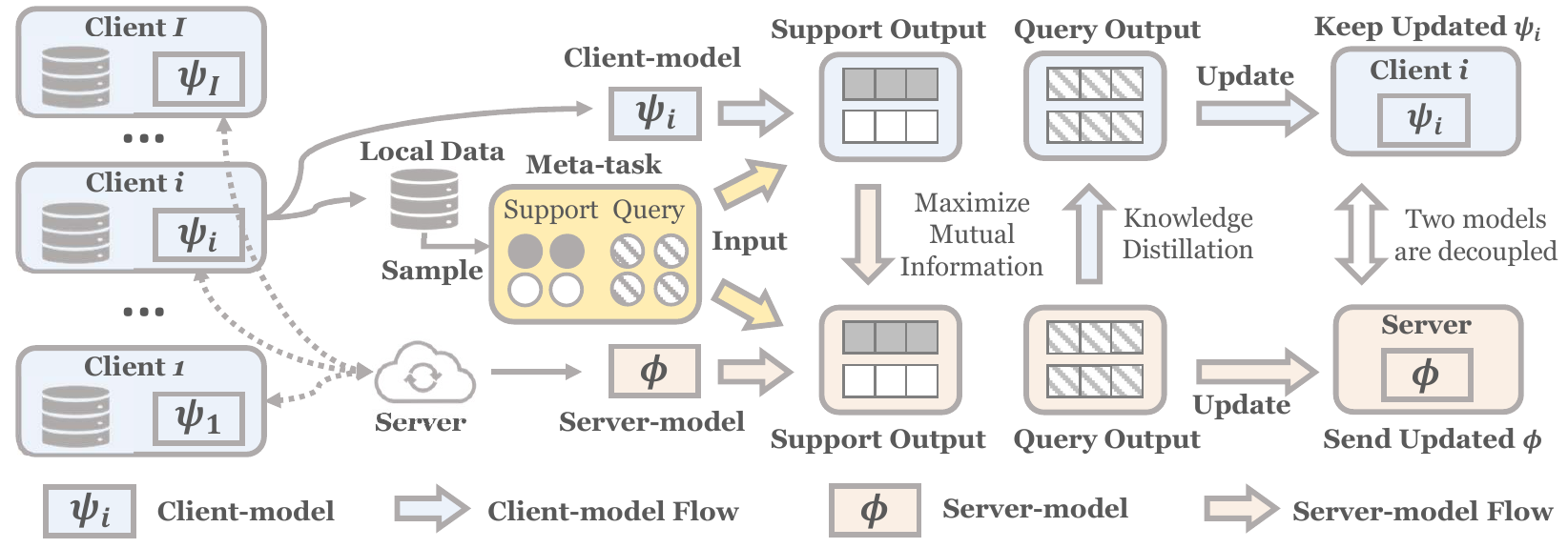}
	   
	    \caption{An illustration of the overall process of our framework F$^2$L. Specifically, each client receives the server-model from the server at the beginning of each round. To perform one step of local update, each client first samples a meta-task (2-way 2-shot in the illustration), which consists of a support set and a query set, from the local data. Then the server-model and the client-model will both compute output for the support samples and query samples. After that, the server-model and the client-model are updated via mutual information maximization and knowledge distillation, respectively. Finally, the server-model is sent back to the server for aggregation, while the client-model is locally preserved by each client.}

	    \label{fig:model}
	    	    \vspace{-0.1in}
	\end{figure*}

In this part, we introduce the overall design of our proposed
framework F$^2$L in detail. 
Specifically, we formulate the \emph{federated few-shot learning} problem under the prevailing $N$-way $K$-shot learning framework. 
Our target of conducting federated few-shot learning is to learn meta-knowledge from a set of $I$ clients $\{\mathbb{C}^{(i)}\}_{i=1}^I$ with different data distributions, and generalize such meta-knowledge to meta-test tasks.
Nevertheless, it remains difficult to conduct federated few-shot learning due to the challenging issues of global data variance and local data insufficiency as mentioned before. Therefore, as illustrated in Fig~\ref{fig:decouple}, we propose a decoupled meta-learning framework to mitigate disruption from the servers. We further propose two update strategies to leverage global knowledge. The overview process is presented in Fig~\ref{fig:model}.

\subsection{Decoupled Meta-Learning Framework}
\label{sec:overall}
\subsubsection{Federated Learning Framework} We consider a server-model, which consists of an encoder $q_\phi$ and a classifier $f_\phi$ that are shared among clients. We denote the overall model parameters in the server-model as $\phi$.
Specifically, $q_\phi:\mathbb{R}^d\rightarrow\mathbb{R}^k$ is a function that maps each sample into a low-dimensional vector $\mathbf{h}_\phi\in\mathbb{R}^k$, where $d$ is the input feature dimension, and $k$ is the dimension of learned representations. Taking the representation $\bh_\phi$ as input, the classifier $f_\phi: \mathbb{R}^k\rightarrow\mathcal{C}_b$ maps each $\bh_\phi$ to the label space of base classes $\mathcal{C}_b$ and outputs the prediction $\bp_\phi\in\mathbb{R}^{|\mcC_b|}$, where each element in $\bp_\phi$ denotes the classification probability regarding each class in $\mathcal{C}_b$.

Following the prevalent FedAvg~\cite{mcmahan2017communication} strategy for FL, the training of server-model is conducted on all clients through $T$ rounds. In each round $t$, the server $\mathbb{S}$ first sends the server-model parameters $\phi$ to all clients, and each client will conduct a local meta-training process on $\tau$ randomly sampled meta-training tasks. Then the server $\mathbb{S}$ will
perform aggregation on parameters received from clients:
	\begin{equation}
	    \phi^{t+1}=\frac{1}{I}\sum\limits_{i=1}^{I}\widetilde{\phi}_i^{t},
     \label{eq:aggregate}
	\end{equation}
    where $\widetilde{\phi}_i^{t}$ denotes the locally updated server-model parameters by client $\mathbb{C}^{(i)}$ on round $t$. $\phi^{t+1}$ denotes the aggregated server-model parameters which will be distributed to clients at the beginning of the next round. In this way, the server can learn a shared model for all clients in a federated manner.

	Although the standard strategy of learning a single shared model for all clients achieves decent performance on general FL tasks~\cite{mcmahan2017communication,fallah2020personalized},
 it can be suboptimal for federated few-shot learning. Due to the global data variance among clients, the aggregated model on the server will disrupt the learning of meta-knowledge in each client~\cite{karimireddy2020scaffold}.
 As a result, the local learning of meta-knowledge in clients
 will become more difficult.
 In contrast, we propose to further introduce a client-model, which is uniquely learned and preserved by each client, to locally learn meta-knowledge. In other words, its model parameters will not be sent back to the server for aggregation. In this manner, we can separate the learning of client-model (meta-knowledge) and server-model (client-invariant knowledge) so that the learning of meta-knowledge is not disrupted.

	
	Specifically, for client $\mathbb{C}^{(i)}$, the client-model also consists of an encoder $q_{\psi_i}$ and a classifier $f_{\psi_i}$. 
 We denote the overall model parameters in the client-model for client $\mathbb{C}^{(i)}$ as $\psi_i$.
In particular, the encoder $q_{\psi_i}$ takes the representation $\bh_\phi$ learned by the encoder $q_{\phi}$ in server-model as input, and outputs a hidden representation $\bh_\psi\in\mathbb{R}^k$. 
Such a design ensures that the client-model encoder $q_{\psi_i}$ does not need to process the raw sample and thus can be a small model, which is important when clients only preserve limited computational resources~\cite{chen2018federated}.
Then the classifier $f_{\psi_i}$ maps $\bh_\psi$ to predictions $\bp_\psi\in\mathbb{R}^N$ of the $N$ classes.


\subsubsection{Local Meta-training on Clients}
Based on the episodic learning strategy, in each round, the training process of each client $\mathbb{C}^{(i)}$ is conducted through $\tau$ steps, where each step is a local update based on a meta-training task randomly sampled from the local base dataset $\mathcal{D}^{(i)}_b$. In particular, for client $\mathbb{C}^{(i)}$ on round $t=1,2,\dotsc,T$ and step $s=1,2,\dotsc,\tau$, we denote the sampled meta-task as $\mathcal{T}_{i}^{t,s}=\{\mathcal{S}_{i}^{t,s},\mathcal{Q}_{i}^{t,s}\}$. 
To learn meta-knowledge from meta-task $\mathcal{T}_{i}^{t,s}$, we adopt the prevalent MAML~\cite{finn2017model} strategy to update client-model in one fine-tuning step and one meta-update step. We first fine-tune the client-model to fast adapt it to support set $\mathcal{S}_{i}^{t,s}$:
	\begin{equation}
	    \widetilde{\psi}_{i}^{t,s}=\psi_{i}^{t,s}-\alpha_{ft}\nabla_\psi\mathcal{L}_{ft}\left(\mathcal{S}_{i}^{t,s};\{\phi_{i}^{t,s},\psi_{i}^{t,s}\}\right),
	   \label{eq:ft}
	   \end{equation}
	where $\mathcal{L}_{ft}$ is the fine-tuning loss, which is the cross-entropy loss calculated on the support set $\mathcal{S}_{i}^{t,s}$. Here, $\alpha_{ft}$ is the learning rate, and $\psi_{i}^{t,s}$ (or $\phi_{i}^{t,s}$) denotes the parameters of client-model (or server-model) on round $t$ and step $s$.
 Then we update the client-model based on the query set $\mathcal{Q}_{i}^{t,s}$: 
	\begin{equation}
	    	    \psi_{i}^{t,s+1}=\psi_{i}^{t,s}-\alpha_\psi\nabla_\psi\mathcal{L}_{\psi}\left(\mathcal{Q}_{i}^{t,s};\{\phi_{i}^{t,s},\widetilde{\psi}_{i}^{t,s}\}\right),
	    	    \label{eq:few_model}
	\end{equation}
where $\mathcal{L}_{\psi}$ is the loss for client-model on the query set $\mathcal{Q}_{i}^{t,s}$, and $\alpha_\psi$ is the meta-learning rate for $\psi$. 
In this regard, we can update client-model with our global-to-local knowledge distillation strategy.
	For the update of server-model, we conduct one step of update based on the support set and parameters of client-model:
	\begin{equation}
	    \phi_{i}^{t,s+1}=\phi_{i}^{t,s}-\alpha_\phi\nabla_{\phi}\mathcal{L}_{\phi}\left(\mathcal{S}_{i}^{t,s};\{\phi_{i}^{t,s}, \psi_{i}^{t,s}\}\right),
	    \label{eq:all_model}
	\end{equation}
	where $\mathcal{L}_{\phi}$ is the loss for the server-model, and $\alpha_\phi$ is the meta-learning rate for $\phi$.
 In this manner, we can update the server-model with our local-to-global knowledge transfer strategy.
 After repeating the above updates for $\tau$ steps, the final parameters of server-model $\phi_i^{t,\tau}$ is used as $\widetilde{\phi}_i^{t}$ in Eq.~(\ref{eq:aggregate}) and sent back to the server for aggregation, while the client-model (with parameters $\psi_i^{t,\tau}$) will be kept locally.
	By doing this, we can decouple the learning of local meta-knowledge in client-model while learning client-invariant knowledge in server-model to avoid disruption from the server.

	%
	%

	\subsection{Local-to-Global Knowledge Transfer}
	With our decoupled meta-learning framework, we can mitigate the disruption to the learning of local meta-knowledge in each client. Nevertheless, we still need to transfer the learned meta-knowledge to server-model (i.e., Local-to-global Knowledge Transfer), so that it can be further leveraged by other clients to handle the local data insufficiency issue.
 Specifically, to effectively transfer local meta-knowledge, we propose to maximize the mutual information between representations learned from server-model encoder $q_\phi$ and client-model encoder $q_\psi$. In this way, the server-model can maximally absorb the information in the learned local meta-knowledge.
	
 \subsubsection{Mutual Information Maximization} Given a meta-training task $\mcT=\{\mathcal{S},\mathcal{Q}\}$, as described in Sec.~\ref{sec:overall}, the server-model encoder $q_\phi$ and client-model encoder $q_\psi$ will output $\mathbf{h}_\phi$ and $\bh_\psi$ for each sample, respectively. By stacking the learned representations of samples in the support set $\mathcal{S}$ ($|\mathcal{S}|=D$, where $D=N \times K$), we can obtain the representations of support samples learned by the server-model, i.e., $\mathbf{H}_\phi\in\mathbb{R}^{D\times k}$, and the client-model, i.e., $\mathbf{H}_\psi\in\mathbb{R}^{D\times k}$.
	For simplicity, we omit the annotations of round $t$, step $s$, and client $i$. 
 The objective of maximizing the information between $\bH_\phi$ and $\bH_\psi$ can be formally represented as follows:
	\begin{equation}
	    	    \max_{\phi} I(\bH_\phi;\bH_\psi)= \max_{\phi}\sum\limits_{i=1}^D\sum\limits_{j=1}^D p(\bh_\phi^i,\bh_\psi^j;\phi)\log\frac{p(\bh_\psi^j|\bh_\phi^i;\phi)}{p(\bh_\phi^j;\phi)},
	\end{equation}
    where 
    $\bh_\phi^i$ (or $\bh_\psi^i$) is 
    the $i$-th row of $\bH_\phi$ (or $\bH_\psi$).
    Since the mutual information $I(\bH_\phi;\bH_\psi)$ is difficult to obtain and thus infeasible to be maximized~\cite{oord2018representation}, we re-write it to achieve a more feasible form:
	\begin{equation}
		    \begin{aligned}
	    I(\bH_\phi;\bH_\psi)
	   	    &=\sum\limits_{i=1}^D\sum\limits_{j=1}^D p(\bh_\phi^i|\bh_\psi^j;\phi)p(\bh_\psi^j;\phi)\log\frac{p(\bh_\psi^j|\bh_\phi^i;\phi)}{p(\bh_\psi^j;\phi)} 
	    .
	    \end{aligned}
	\end{equation}
	Since the support set $\mathcal{S}$ of size $D$ is randomly sampled, we can assume that the prior probability $p(\bh_\psi^j;\phi)$ follows a uniform distribution,	and set it as $p(\bh_\psi^j;\phi)=1/D$. According to the Bayes' theorem, the Eq. (7) becomes:
		\begin{equation}
		    \begin{aligned}
	    I(\bH_\phi;\bH_\psi)
	    &=\frac{1}{D}\sum\limits_{i=1}^D\sum\limits_{j=1}^D p(\bh_\phi^i|\bh_\psi^j;\phi)\left(\log(p(\bh_\psi^j|\bh_\phi^i;\phi))+\log D\right)
	    .
	    \label{eq:objective}
	    \end{aligned}
	\end{equation}
We next present alternative strategies to estimate $p(\bh_\phi^i|\bh_\psi^j;\phi)$ and $p(\bh_\psi^j|\bh_\phi^i;\phi)$ in detail.
	 \subsubsection{Estimation of $p(\bh_\phi^i|\bh_\psi^j;\phi)$} 
  Since the client-model is fine-tuned on the support set $\mathcal{S}$ of the meta-task $\mathcal{T}$, we can leverage the classification results of the client-model to estimate $p(\bh_\phi^i|\bh_\psi^j;\phi)$. We denote $C(j)$ as the set of sample indices in the support set $\mathcal{S}$ that shares the same class as the $j$-th sample (including itself), i.e., $C(j)\equiv\{k:y_k=y_j,k=1,2,\dotsc,D\}$. Here, we first set $p(\bh_\phi^i|\bh_\psi^j;\phi)=0$ for all $i\notin C(j)$, since we assume the client-model can only infer representations from the same class.
	Intuitively, in the case of $i\in C(j)$, which means the $i$-th and the $j$-th samples share the same class, $p(\bh_\phi^i|\bh_\psi^j;\phi)$ can be considered as the confidence of client-model regarding the class of the $j$-the sample. Therefore, it should reflect the degree to which the sample representation $\bh_\psi^j$ is relevant to its class. Utilizing the client-model classification output (i.e., normalized class probabilities) for the $i$-th sample $\bp^i_\psi\in\mathbb{R}^N$, we can compute $p(\bh_\phi^i|\bh_\psi^j;\phi)$ as follows:
 		\begin{equation}
	    p(\bh_\phi^i|\bh_\psi^j;\phi)=\left\{\begin{aligned}\frac{\bp^i_\psi(y_j)}{\sum_{k\in C(j)}\bp^k_\psi(y_j)}&\ \ \ \ \ \ \text{if}\ \  i\in C(j)\\0 \ \ \ \ \ \ \ \ \ \ \  &
	    \ \ \ \ \ \ \text{otherwise}\end{aligned}\right.
	    ,
	\end{equation}
	where $\bp_\psi^i(y_j)\in\mathbb{R}$ denotes the classification probability for the $i$-th sample regarding class $y_j$ ($y_i=y_j$ when $i\in C(j)$). 
 
 \subsubsection{Estimation of $p(\bh_\psi^j|\bh_\phi^i;\phi)$}  Next we elaborate on how to estimate $p(\bh_\psi^j|\bh_\phi^i;\phi)$. Although we can similarly leverage the classification results of the server-model, such a strategy lacks generalizability. This is because the server-model aims at classifying all base classes instead of the 
 $N$ classes in each meta-training task.
 We instead propose to estimate $p(\bh_\psi^j|\bh_\phi^i;\phi)$ based on the Euclidean distance (divided by 2 for simplicity) between learned representations of the server-model and the client-model. Specifically, we normalize the distances
 with a softmax function:
	    \begin{equation}
        \left.p(\bh_\psi^j|\bh_\phi^i;\phi)
        =\frac{\exp\left(-\|\bh_\phi^i- \bh_\psi^j\|_2^2/2\right)}{\sum_{k\in C(i)}\exp\left(-\|\bh_\phi^i- \bh_\psi^k\|_2^2/2\right)}
        \right.
        .
    \end{equation}
    Then if we further apply the $\ell_2$ normalization to both $\bh_\phi^i$ and $\bh_\psi^j$, we can obtain $\|\bh_\phi^i- \bh_\psi^j\|_2^2/2=1-\bh_\phi^i\cdot \bh_\psi^j$. Moreover, since the value of $\sum_{i=1}^D\sum_{j=1}^Dp(\bh_\phi^i|\bh_\psi^j;\phi)$ equals a constant $D$, the  term $\sum_{i=1}^D\sum_{j=1}^Dp(\bh_\phi^i|\bh_\psi^j;\phi)\cdot\log(D)/D$ in Eq. (\ref{eq:objective}) is also a constant and thus can be ignored in the objective:
    \begin{equation}
        \frac{1}{D}\sum\limits_{i=1}^D\sum\limits_{j=1}^D p(\bh_\phi^i|\bh_\psi^j;\phi)\log\left(D\right)=\frac{1}{D}\cdot D\cdot\log\left(D\right)=\log\left(D\right).
    \end{equation}
    Combining the above equations, the optimal server-model parameter $\phi^*$ for the final optimization objective (i.e., $\max_{\phi} I(\bH_\phi;\bH_\psi)$) can be obtained as follows:
    \begin{equation}
 \phi^*=\argmax_{\phi}I(\bH_\phi;\bH_\psi)=\argmin_{\phi}\mathcal{L}_{MI}.
    \end{equation}
    Here $\mathcal{L}_{MI}$ is defined as follows:
        \begin{equation}
        \begin{aligned}
\mathcal{L}_{MI}
&=\frac{1}{D}\sum\limits_{i=1}^D\sum\limits_{j=1}^D p(\bh_\phi^i|\bh_\psi^j;\phi)\log(p(\bh_\psi^j|\bh_\phi^i;\phi))\\
&=\frac{1}{D} \sum\limits_{j=1}^D\sum\limits_{i\in C(j)}
        -\frac{\bp^i_\psi(y_j)\left(\bh_\phi^i\cdot\bh_\psi^j\right)}{\sum_{k\in C(j)}\bp^k_\psi(y_j)}
        \\
        &+\frac{\bp^i_\psi(y_j)}{\sum_{k\in C(j)}\bp^k_\psi(y_j)}
        \log\left(\sum_{k\in C(i)}\exp\left(\bh_\phi^i\cdot \bh_\psi^k\right)\right)
        ,
        \end{aligned}
    \label{eq:mi_objective}
    \end{equation}
    where we exchange the order of summation over $i$ and $j$ for clarity. 
    It is noteworthy that $\mathcal{L}_{MI}$ is different from the InfoNCE loss~\cite{oord2018representation,he2020momentum}, which considers different augmentations of samples, while $\mathcal{L}_{MI}$ focuses on the classes of samples in $\mathcal{S}$.
    Moreover, $\mathcal{L}_{MI}$ also differs from the supervised contrastive loss~\cite{khosla2020supervised}, which combines various augmentations of samples and label information. In contrast, our loss targets at transferring the meta-knowledge by maximally preserving the mutual information between representations learned by the server-model and the client-model.
    More differently, the term $\bp^i_\psi(y_j)/\sum_{k\in C(j)}\bp^k_\psi(y_j)$ acts as an adjustable weight that measures the importance of a sample to its class. Combining the objective described in Eq.~(\ref{eq:mi_objective}) and the standard cross-entropy loss, we can obtain the final loss for the server-model:
	    \begin{equation}
    \begin{aligned}
    \mathcal{L}_{\phi}
    &=(1-\lambda_{MI})\mathcal{L}_{CE}(\mathcal{S})+\lambda_{MI}\mathcal{L}_{MI},
        \end{aligned}
    \label{eq:transfer_objective}
    \end{equation}
where $\mathcal{L}_{CE}(\mathcal{S})$ is defined as follows:
    	    \begin{equation}
    \begin{aligned}
\mathcal{L}_{CE}(\mathcal{S})
    &=-\frac{1}{D}\sum\limits_{i=1}^D\sum\limits_{j=1}^{|\mathcal{C}_b|}y^i_{c_j}\log \bp^i_\phi(c_j),
        \end{aligned}
    \end{equation}
	where $\bp_\phi^i(c_j)\in\mathbb{R}$ denotes the classification probability for the $i$-th support sample belonging to the $j$-th class $c_j$ in $\mathcal{C}_b$, computed by the server-model. Here $y^i_{c_j}=1$ if the $i$-th support sample belongs to $c_j$, and $y^i_{c_j}=0$, otherwise. Moreover, $\lambda_{MI}\in[0,1]$ is an adjustable hyper-parameter to control the weight of $\mathcal{L}_{MI}$.

	\subsection{Global-to-Local Knowledge Distillation}
With the learned meta-knowledge in each client transferred from the client-model to the server-model, 
other clients can leverage such meta-knowledge to deal with the local data insufficiency issue. 
However, since each meta-task only contains $N$ classes, directly extracting meta-knowledge in the server-model can inevitably involve meta-knowledge from other classes, which can be harmful to the learning of local meta-knowledge from these $N$ classes in each client.
Instead, we propose a partial knowledge distillation strategy to selectively extract useful knowledge from the server-model, i.e., global-to-local knowledge distillation.
 
 \subsubsection{Partial Knowledge Distillation} Specifically, we focus on the output classification probabilities of the server-model regarding the $N$ classes in support set $\mathcal{S}$ while ignoring other classes. In this regard, we can extract the information that is crucial for learning local meta-knowledge from these $N$ classes and also reduce the irrelevant information from other classes.
	
	
	Particularly, we consider the same meta-task $\mathcal{T}=\{\mathcal{S},\mathcal{Q}\}$.
 We denote the output probabilities for the $i$-th query sample $q_i$ in $\mathcal{Q}$ (with label $y_i$) of the server-model and the client-model as $\bp^i_\phi\in\mathbb{R}^{|\mathcal{C}_b|}$ and $\bp^i_\psi\in\mathbb{R}^{N}$, respectively.
 It is noteworthy that the $N$ classes in this meta-task, denoted as $\mathcal{C}_m$, are sampled from the base classes $\mathcal{C}_b$ (i.e., $|\mathcal{C}_m|=N$ and $\mathcal{C}_m\subset\mathcal{C}_b$). Therefore, the output of server-model (i.e., $\bp^i_\phi$) will include the probabilities of classes in $\mathcal{C}_m$. In particular, we enforce the probabilities of 
 in $\mcC_m$ from the client-model to be consistent with the probabilities of the same classes from the server-model. As a result, the learning of local meta-knowledge can leverage the 
 information of data in the same $N$ classes from other clients, which is encoded in the server-model.
 In this regard, we can handle the local data insufficiency issue by involving information from other clients while reducing the irrelevant information from other classes not in $\mathcal{C}_m$. In particular, by
 utilizing the output of the server-model as the soft target for the client-model, we can achieve an objective 
 as follows:
	\begin{equation}
	    	\mathcal{L}_{KD}=-\frac{1}{Q}\sum\limits_{i=1}^Q\sum\limits_{j=1}^N\bq^i_\phi(c_j)\log\bq^i_\psi(c_j),
	\end{equation}
    where $c_j$ is the $j$-th class in $\mcC_m$ (i.e., the $N$ classes in meta-task $\mcT$). $\bq^i_\phi(c_j)$ and $\bq^i_\psi(c_j)$ are the knowledge distillation values for $c_j$ from server-model and client-model, respectively. Specifically, the values of $\bq^i_\phi(c_j)$ and $\bq^i_\psi(c_j)$ are obtained via the softmax normalization:
    \begin{equation}
        \bq^i_\phi(c_j)=\frac{\exp(\bz_\phi^i(c_j)/T_i)}{\sum_{k=1}^N\exp(\bz_\phi^i(c_k)/T_i)},
    \end{equation}
	    \begin{equation}
        \bq^i_\psi(c_j)=\frac{\exp(\bz_\psi^i(c_j)/T_i)}{\sum_{k=1}^N\exp(\bz_\psi^i(c_k)/T_i)},
    \end{equation}
	where $\bz_\phi^i(c_j)$ are $\bz_\psi^i(c_j)$) are the logits (i.e., output before softmax normalization) of class $c_j$ from server-model and client-model, respectively. $T_i$ is the temperature parameter for the $i$-th query sample. In this way, we can ensure that $\sum_{j=1}^N\bq_\phi^i(c_j)=\sum_{j=1}^N\bq_\psi^i(c_j)=1$. 
 
	\subsubsection{Adaptive Temperature Parameter} Generally, a larger value of $T_i$ denotes that the client-model focuses more on extracting information from the other classes in $\mcC_m$~\cite{hinton2015distilling} (i.e., $\{c|c\in\mcC_m,c\neq y_i\}$), denoted as negative classes. Since the classification results can be erroneous in the server-model, we should adaptively adjust the value of $T_i$ for each meta-task to reduce the adverse impact of extracting misleading information from the server-model. 
	However, although negative classes can inherit useful information for classification, such information is generally noisier when the output probabilities of these negative classes are smaller. Therefore, to estimate the importance degree of each negative class, we consider the maximum output logit for negative classes to reduce potential noise. 
	Particularly, if the probability of a negative class from the server-model is significantly larger than other classes, we can conjecture that this class is similar to $y_i$ and thus potentially contains the crucial information to distinguish them.
	Specifically, the temperature parameter $T_i$ for the $i$-th query sample $q_i$ is computed as follows:
	\begin{equation}
	    T_i=\sigma\left(\frac{\max_{c\in\mathcal{C}_{m},c\neq y_i}\exp(\bz_\phi^i(c))}{\exp(\bz_\phi^i(y_i))}\right),
	\end{equation}
	where $\sigma(\cdot)$ denotes the Sigmoid function, and $y_i$ is the label of $q_i$.
	In this way, the temperature parameter $T_i$ will increase when the ratio between the largest probability in negative classes and the probability for $y_i$ is larger. As a result, the client-model will focus more on the negative class information.
 Then by further incorporating the cross-entropy loss on the query set $\mathcal{Q}$, we can obtain the final loss for the client-model:
	\begin{equation}
	    \mathcal{L}_{\psi}=(1-\lambda_{KD})\mathcal{L}_{CE}(\mathcal{Q})+\lambda_{KD}\mathcal{L}_{KD},
	    	    \label{eq:achieve_objective}
	\end{equation}
 where $\mathcal{L}_{CE}(\mathcal{Q})$ is defined as follows:

	\begin{equation}
	   \mathcal{L}_{CE}(\mathcal{Q})=-\frac{1}{Q}\sum\limits_{i=1}^Q\sum\limits_{j=1}^Ny^i_{c_j}\log \bp^i_\psi(c_j),
	\end{equation}
	where $\bp^i_\psi(c_j)$ is the probability of the $i$-th query sample belonging to class $c_j$ computed by the client-model.
	$y^i_{c_j}=1$ if the $i$-th query sample belongs to $c_j$, and $y^i_{c_j}=0$, otherwise. Moreover, $\lambda_{KD}\in[0,1]$ is an adjustable hyper-parameter to control the weight of $\mathcal{L}_{KD}$. In this manner, the client-model can selectively learn useful knowledge from both the local and global perspectives, i.e., global-to-local knowledge distillation.


	

	\subsection{Overall Learning Process}
        With the 
        proposed losses $\mathcal{L}_\phi$ and $\mathcal{L}_\psi$, on each round, we can conduct meta-training on each client $\mathbb{C}^{(i)}$ by sampling $\tau$ meta-training tasks from the local base dataset $\mathcal{D}^{(i)}_b$. The detailed process is described in Algorithm~\ref{algorithm}.
	After $T$ rounds of meta-training on all the clients, we have obtained a model that accommodates comprehensive meta-knowledge for federated few-shot learning. For the meta-test phase, 
 since we have aggregated learned local meta-knowledge from each client to the server-model, we can leverage the server-model to generate data representations for classification. Specifically, during evaluation, for each meta-test task $\mathcal{T}=\{\mathcal{S},\mathcal{Q}\}$ sampled from local novel datasets $\{\mathcal{D}_n^{(i)}\}_{i=1}^I$ in all clients, 
 we follow the same process as meta-training including fine-tuning, except that the meta-update process is omitted. The output of the client-model will be used for classification.
	
	
				   					\begin{table*}[htbp]
			   							   					\setlength\tabcolsep{5.2pt}
		\centering
		\renewcommand{\arraystretch}{1.2}

			\caption{The overall federated few-shot learning results of various models on four datasets under IID and Non-IID settings (5-way), where accuracy and standard deviation are reported in $\%$. The best results are presented as \textbf{bold}.}
     \vspace{-0.05in}
	\begin{tabular}{c|c|c|c|c||c|c|c|c}
	\hline
	     Dataset& \multicolumn{4}{c||}{20 Newsgroup}&\multicolumn{4}{c}{Huffpost}\\\hline
	     Distribution&\multicolumn{2}{c|}{IID}&\multicolumn{2}{c||}{Non-IID}&\multicolumn{2}{c|}{IID}&\multicolumn{2}{c}{Non-IID}\\\hline
	    Setting & 1-shot&5-shot&1-shot&5-shot& 1-shot&5-shot& 1-shot&5-shot\\\hline\hline
Local&$31.53\pm1.68$&$42.73\pm1.51$&$29.64\pm1.81$&$41.01\pm2.40$&$34.02\pm1.67$&$49.95\pm1.54$&$33.09\pm2.28$&$47.18\pm1.43$\\\hline
FL-MAML&$32.89\pm1.86$&$44.34\pm1.66$&$31.60\pm1.44$&$43.84\pm1.97$&$37.47\pm1.43$&$52.85\pm1.43$&$36.01\pm2.17$&$50.56\pm2.08$\\\hline
FL-Proto&$35.62\pm2.07$&$46.04\pm1.92$&$32.79\pm1.41$&$43.82\pm1.85$&$37.87\pm1.23$&$51.90\pm1.43$&$34.05\pm1.35$&$50.52\pm1.33$\\\hline
FedFSL&$36.56\pm1.41$&$46.37\pm1.82$&$35.84\pm1.49$&$45.89\pm1.72$&$39.18\pm1.42$&$53.81\pm1.36$&$37.86\pm1.46$&$52.18\pm1.82$\\\hline
F$^2$L&$\mathbf{39.80\pm1.80}$&$\mathbf{49.64\pm1.32}$&$\mathbf{39.00\pm1.36}$&$\mathbf{49.44\pm1.98}$&$\mathbf{42.12\pm2.12}$&$\mathbf{57.88\pm2.17}$&$\mathbf{41.64\pm1.81}$&$\mathbf{57.12\pm1.87}$\\\hline	    
	\end{tabular}

			\label{tab:all_result_text}
	\end{table*}
		   					\begin{table*}[htbp]
		   					\setlength\tabcolsep{5.2pt}
		\centering
		\renewcommand{\arraystretch}{1.2}
	\begin{tabular}{c|c|c|c|c||c|c|c|c}
	\hline
	     Dataset& \multicolumn{4}{c||}{FC100}&\multicolumn{4}{c}{miniImageNet}\\\hline
	     Distribution&\multicolumn{2}{c|}{IID}&\multicolumn{2}{c||}{Non-IID}&\multicolumn{2}{c|}{IID}&\multicolumn{2}{c}{Non-IID}\\\hline
	    Setting & 1-shot&5-shot&1-shot&5-shot& 1-shot&5-shot& 1-shot&5-shot\\\hline\hline
     Local&$33.45\pm1.68$&$50.89\pm1.56$&$32.40\pm1.76$&$50.29\pm2.24$&$47.82\pm1.68$&$64.30\pm1.59$&$46.81\pm2.03$&$64.06\pm1.45$\\\hline
FL-MAML&$34.10\pm1.29$&$50.66\pm1.68$&$36.06\pm1.78$&$50.35\pm1.57$&$49.74\pm1.40$&$65.55\pm1.57$&$47.64\pm1.36$&$63.56\pm1.13$\\\hline
FL-Proto&$36.11\pm1.49$&$54.74\pm2.05$&$35.54\pm1.71$&$52.31\pm1.76$&$51.32\pm1.41$&$66.67\pm2.06$&$50.82\pm1.82$&$65.09\pm1.90$\\\hline
FedFSL&$39.38\pm1.95$&$52.25\pm1.84$&$38.60\pm2.00$&$53.90\pm1.80$&$55.75\pm2.06$&$70.59\pm1.97$&$53.52\pm2.01$&$69.56\pm1.86$\\\hline
F$^2$L&$\mathbf{42.52\pm2.06}$&$\mathbf{58.60\pm2.09}$&$\mathbf{42.56\pm2.25}$&$\mathbf{59.52\pm2.14}$&$\mathbf{56.72\pm1.79}$&$\mathbf{74.23\pm2.32}$&$\mathbf{56.16\pm2.05}$&$\mathbf{73.24\pm2.02}$\\\hline
	    
	\end{tabular}
\vspace{-0.05in}
			\label{tab:all_result_image}
	\end{table*}
	
	\section{Experiments}
	    In this part, we conduct extensive experiments to evaluate our framework F$^2$L on four few-shot classification datasets covering both news articles and images under the federated scenario. 
	\subsection{Datasets}
	In this section, we introduce four prevalent real-world datasets used in our experiments, covering both news articles and images: \textbf{20 Newsgroup}~\cite{lang1995newsweeder}, \textbf{Huffpost}~\cite{misra2018news,misra2021sculpting}, \textbf{FC100}~\cite{oreshkin2018tadam}, and \textbf{miniImageNet}~\cite{vinyals2016matching}. In particular, 20 Newsgroup and Huffpost are online news article datasets, while FC100 and miniImageNet are image datasets. The details are as follows:
	\begin{itemize}
		    \item \textbf{20 Newsgroup}~\cite{lang1995newsweeder} is a text dataset that consists of informal discourse from news discussion forums. There are 20 classes for documents in this dataset, where each class belongs to one of six top-level categories. The classes are split as 8/5/7 for training/validation/test, respectively.
	    \item \textbf{Huffpost}~\cite{misra2018news,misra2021sculpting} is a text dataset containing news headlines published on HuffPost\footnote{https://www.huffpost.com/} between 2012 and
2018. Generally, the headlines are significantly shorter and less grammatical
than the 20 Newsgroup dataset. Moreover, each headline belongs to one of 41 classes, which are then split as 20/5/16 for training/validation/test, respectively.
	    \item \textbf{FC100}~\cite{oreshkin2018tadam} is an image classification dataset based on CIFAR-100~\cite{krizhevsky2009learning}. Specifically, this dataset contains 100 image classes, where each class maintains 600 images with a low $32\times32$ resolution. The classes are split as 60/20/20 for training/validation/test, respectively.
	    \item \textbf{miniImageNet}~\cite{vinyals2016matching} is an image dataset extracted from the full ImageNet dataset~\cite{deng2009imagenet}. This dataset consists of 100 image classes, and each class maintains 600 images with a resolution of $84\times84$. The classes are split as 64/16/20 for training/validation/test, respectively.
	\end{itemize}

\subsection{Experimental Settings}
To validate the performance of our framework \textsc{F$^2$L}, we conduct experiments with the following baselines for a fair comparison:
	\begin{itemize}
    \item  \emph{Local}. This baseline is non-distributed, which means we train an individual model for each client on the local data. The meta-test process is conducted on all meta-test tasks, and the averaged results of all models are reported.
    \item \emph{FL-MAML}. This baseline leverages the MAML~\cite{finn2017model} strategy to perform meta-learning on each client. The updated model parameters will be sent back to the server for aggregation.
    \item \emph{FL-Proto}. This baseline uses ProtoNet~\cite{snell2017prototypical} as the model in each client. The classification is based on the Euclidean distances between query samples and support samples.
    \item \emph{FedFSL}~\cite{fan2021federated}. This method combines MAML and an adversarial learning strategy~\cite{goodfellow2014generative,saito2018maximum} to construct a consistent feature space. The aggregation is based on FedAvg~\cite{mcmahan2017communication}.
 
\end{itemize}
During meta-training, we perform updates for the client-model and the server-model according to Algorithm~\ref{algorithm}. Finally, the server-model that achieves the best result on validation will be used for meta-test. Then during meta-test, we evaluate the server-model on a series of 100 randomly sampled meta-test tasks from local novel datasets $\{\mathcal{D}_n^{(i)}\}_{i=1}^I$ in all clients. For consistency, the class split of $\mathcal{C}_b$ and $\mathcal{C}_n$ is identical for all baseline methods. The classification accuracy over these meta-test tasks will be averaged as the final results. The specific parameter settings are provided in Appendix~\ref{app:parameter}. 
For the specific choices for the encoder and classifier in server-model and client-model (i.e., $q_\phi$, $f_\phi$, $q_\psi$, and $f_\psi$) and model parameters, we provide further details in Appendix~\ref{app:encoder}. Note that for a fair comparison, we utilize the same encoder for all methods. 

\subsection{Overall Evaluation Results}
We present the overall performance comparison of our framework and baselines on federated few-shot learning in Table~\ref{tab:all_result_text}. Specifically, 
we conduct experiments under two few-shot settings: 5-way 1-shot and 5-way 5-shot. Moreover, to demonstrate the robustness of our framework under different data distributions, we partition the data in both IID and non-IID settings. For the IID partition, the samples of each class are uniformly distributed to all clients. For non-IID partition, we follow the prevailing strategy~\cite{hsu2019measuring,yu2020salvaging} and distribute samples to all clients based on the Dirichlet distribution with its concentration parameter set as 1.0. The evaluation metric is the average classification accuracy over ten repetitions. From the overall results, we can obtain the following observations:
\begin{itemize}
    \item Our framework F$^2$L outperforms all other baselines on various news article and image datasets under different few-shot settings (1-shot and 5-shot) and data distributions (IID and non-IID). The results validate the effectiveness of our framework on federated few-shot learning.
   \item Conventional few-shot methods such as Prototypical Network~\cite{snell2017prototypical} and MAML~\cite{finn2017model} exhibit similar performance compared with the Local baseline. The result demonstrates that directly applying few-shot methods to federated learning brings less competitive improvements over local training. This is because such methods are not proposed for federated learning and thus lead to unsatisfactory training performance under the federated setting.
   \item The performance of all methods degrades at different extents when the data distribution is changed from IID to non-IID. The main reason is that the variety of classes in each client results in a more complex class distribution and brings difficulties to the classification task. Nevertheless, by effectively transferring the meta-knowledge among clients, our framework is capable of alleviating such a problem under the non-IID scenario.
   \item When increasing the value of $K$ (i.e., more support samples in each class), all methods achieve considerable performance gains. In particular, our framework F$^2$L obtains better results compared to other baselines, due to our decoupled meta-learning framework, which promotes the learning of meta-knowledge in the support samples.
\end{itemize}
			\begin{figure}[!t]
		\centering
\captionsetup[sub]{skip=-1pt}
\subcaptionbox{FC100}
{\includegraphics[width=0.23\textwidth]{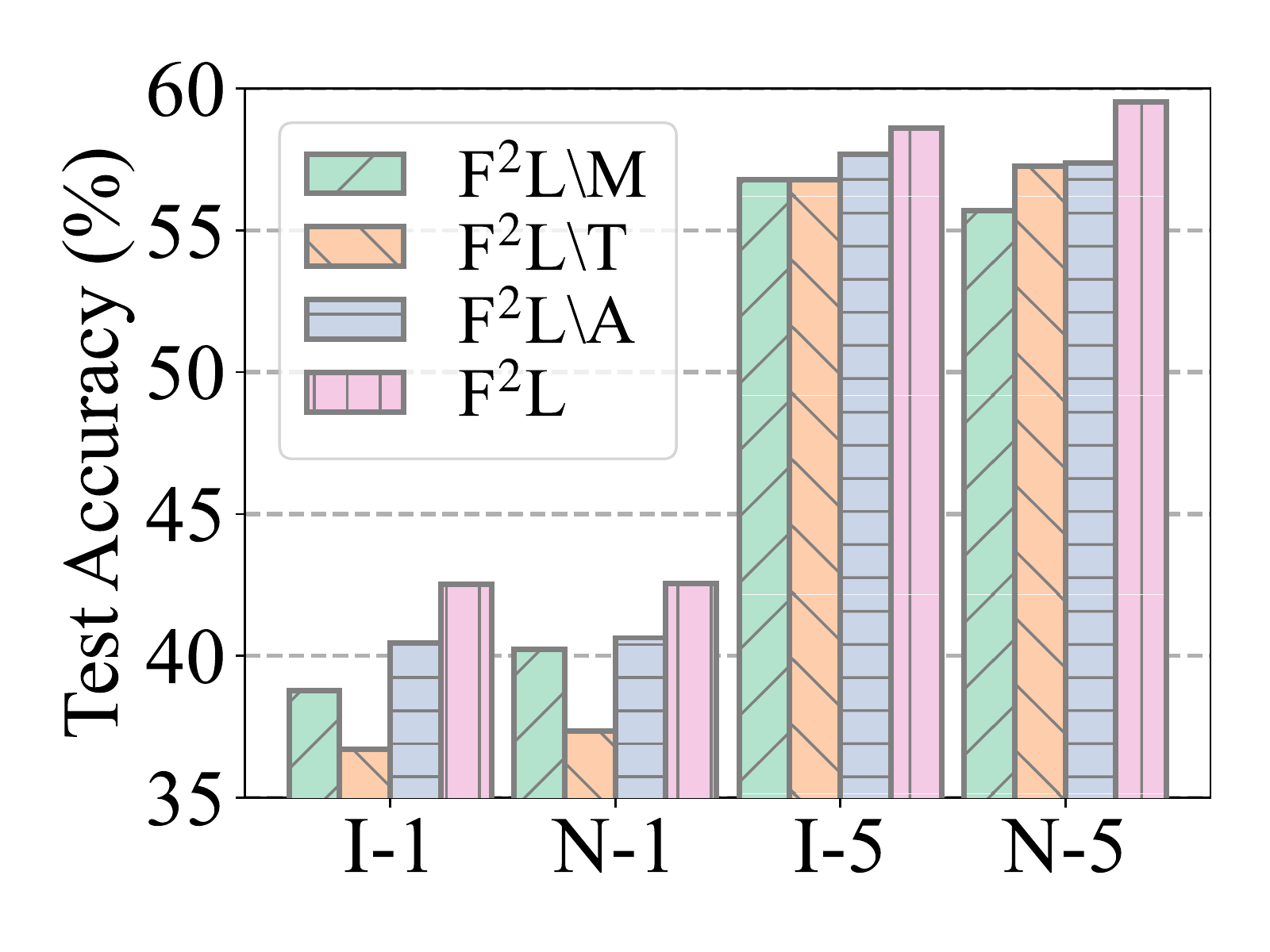}}
\subcaptionbox{Huffpost}
{\includegraphics[width=0.23\textwidth]{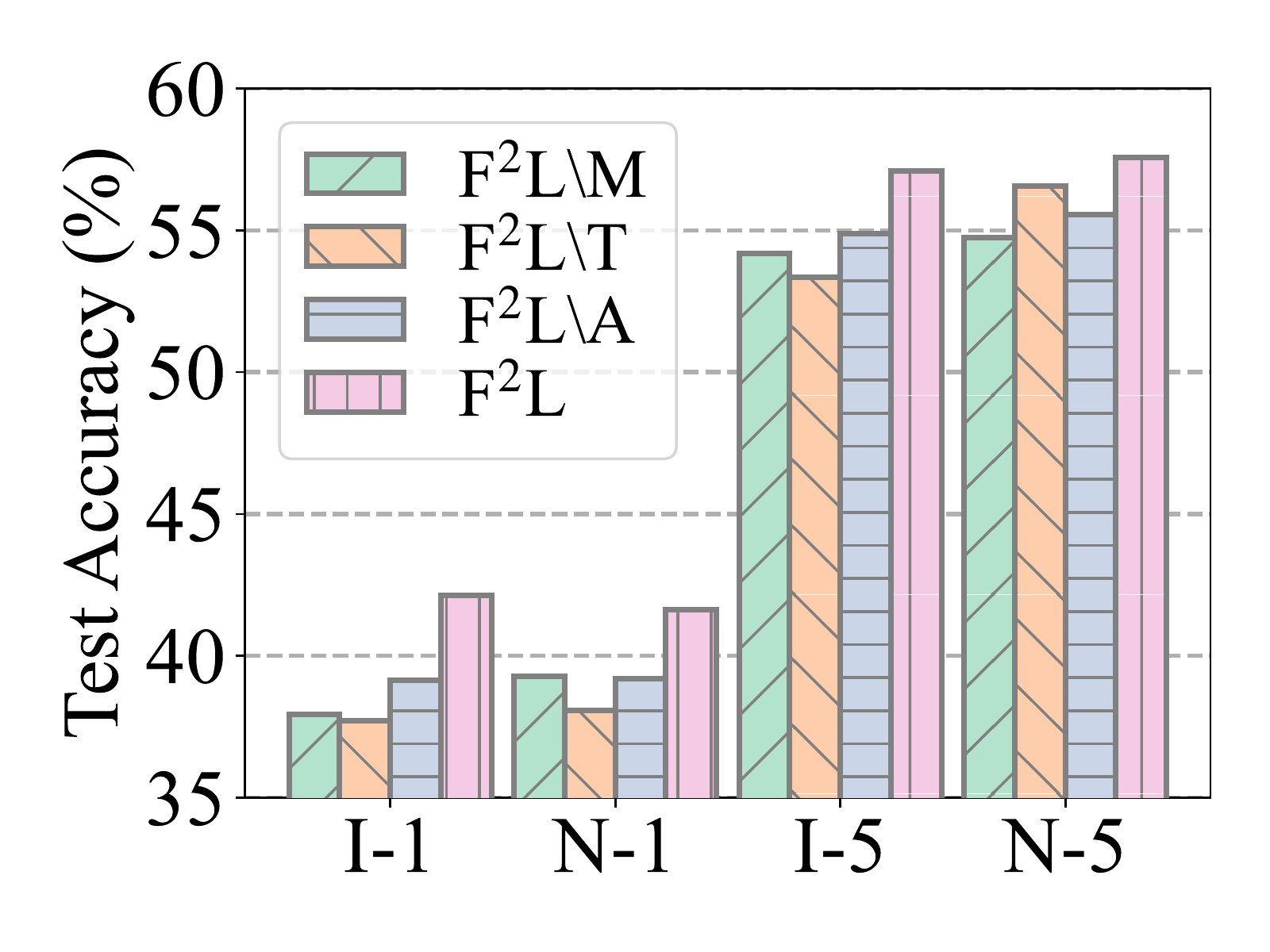}}
\vspace{-0.1in}
		\caption{Ablation study of our framework on FC100 and Huffpost. I-$K$ (or N-$K$) denotes the setting of 5-way $K$-shot under IID (or non-IID) distributions. M denotes the decoupled framework, T means the local-to-global knowledge transfer, and A demotes the global-to-local knowledge distillation. }
		\label{fig:ablation}
\vspace{-0.1in}
	\end{figure}
	
\subsection{Ablation Study}

In this part, we conduct an ablation study on FC100 and Huffpost to validate the effectiveness of three crucial designs in F$^2$L (similar results observed in other datasets). First, we remove the decoupled strategy so that the client-model will also be sent to the server for aggregation. 
We refer to this variant as \emph{F$^2$L\textbackslash M}. Second, we remove the local-to-global knowledge transfer module so that the meta-knowledge in the client-model will be effectively transferred to the server-model. This variant is referred to as \emph{F$^2$L\textbackslash T}. Third, we eliminate the global-to-local knowledge distillation loss. In this way, the client-model cannot leverage the global knowledge in the server-model for learning meta-knowledge. We refer to this variant as \emph{F$^2$L\textbackslash A}. The overall ablation study results are presented in Fig.~\ref{fig:ablation}.
From the results, we observe that F$^2$L outperforms all variants, which verifies the effectiveness of the three designs in F$^2$L. Specifically, removing the design of local-to-global knowledge transfer leads to significant performance degradation. This result demonstrates that such a design can effectively aggregate learned meta-knowledge among clients and thus bring performance improvements. 
More significantly, without our decoupled strategy, the performance deteriorates rapidly when federated few-shot learning is conducted in the non-IID scenario. This phenomenon verifies the importance of mitigating the disruption from the server in the presence of complex data distributions among clients.

			\begin{figure}[!t]
		\centering
\captionsetup[sub]{skip=-1pt}
\subcaptionbox{1-shot}
{\includegraphics[width=0.23\textwidth]{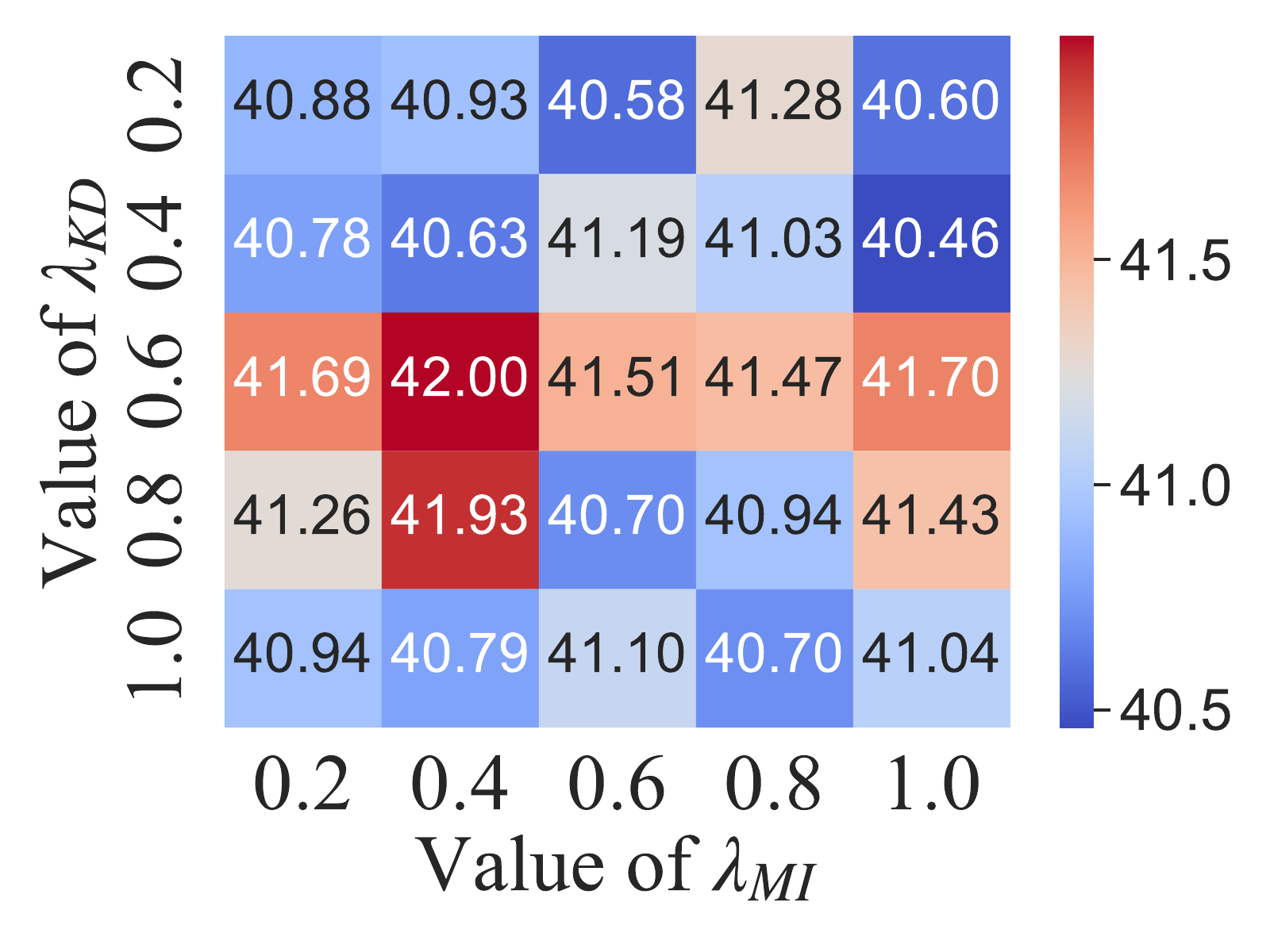}}
\subcaptionbox{5-shot}
{\includegraphics[width=0.23\textwidth]{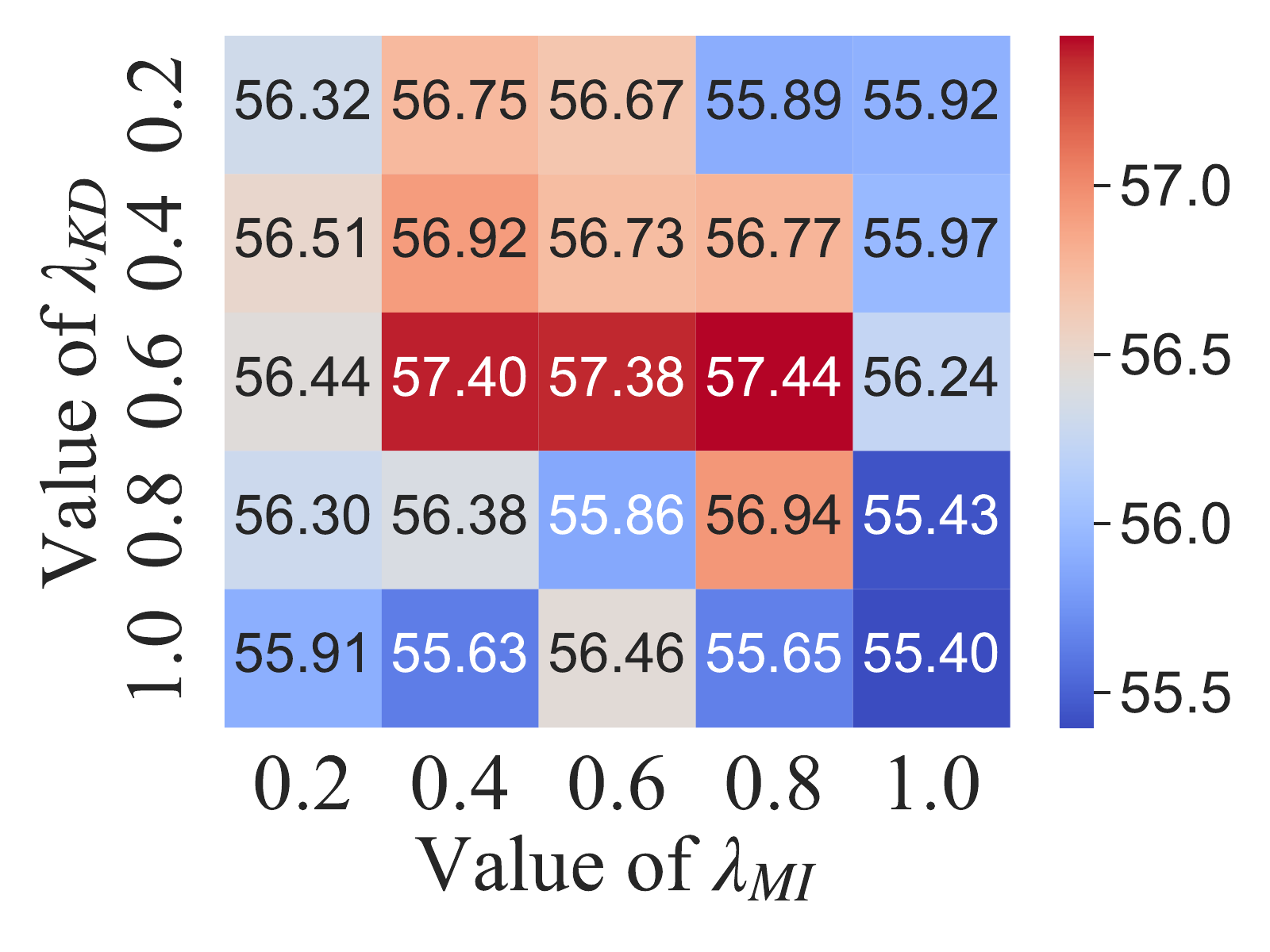}}
\vspace{-0.1in}
		\caption{The results with different values of $\lambda_{MI}$ and $\lambda_{KD}$ on Huffpost under the non-IID setting.}
		\label{fig:loss}
\vspace{-0.15in}
	\end{figure}
	
\subsection{Parameter Sensitivity Study}
\subsubsection{Effect of $\lambda_{MI}$ and $\lambda_{KD}$}
In this section, we further conduct experiments to study the sensitivity of several parameters in our framework F$^2$L. During the process of transferring and achieving meta-knowledge, we introduce two novel losses $\mathcal{L}_{MI}$ and $\mathcal{L}_{KD}$, respectively, along with the traditional cross-entropy loss. 
To empirically evaluate the impact brought by different values of $\lambda_{MI}$ and $\lambda_{KD}$ in Eq. (\ref{eq:transfer_objective}) and Eq. (\ref{eq:achieve_objective}) , we adjust the values of $\lambda_{MI}$ and $\lambda_{KD}$ from $0$ to $1$ and present the results in Fig~\ref{fig:loss}. From the results, we can observe that the performance generally increases with a larger value of $\lambda_{MI}$, while decreasing with $\lambda_{MI}$ approaches 1. The results indicate the importance of transferring learned local meta-knowledge, while also demonstrating that the cross-entropy loss is necessary. On the other hand, the performance first increases and then degrades when a larger value of $\lambda_{KD}$ is presented. That being said, although partial knowledge distillation can enable each client to benefit from the global data, a larger $\lambda_{KD}$ can potentially lead to more irrelevant information when learning local meta-knowledge.

			\begin{figure}[!t]
		\centering
\captionsetup[sub]{skip=-1pt}
\subcaptionbox{1-shot}
{\includegraphics[width=0.23\textwidth]{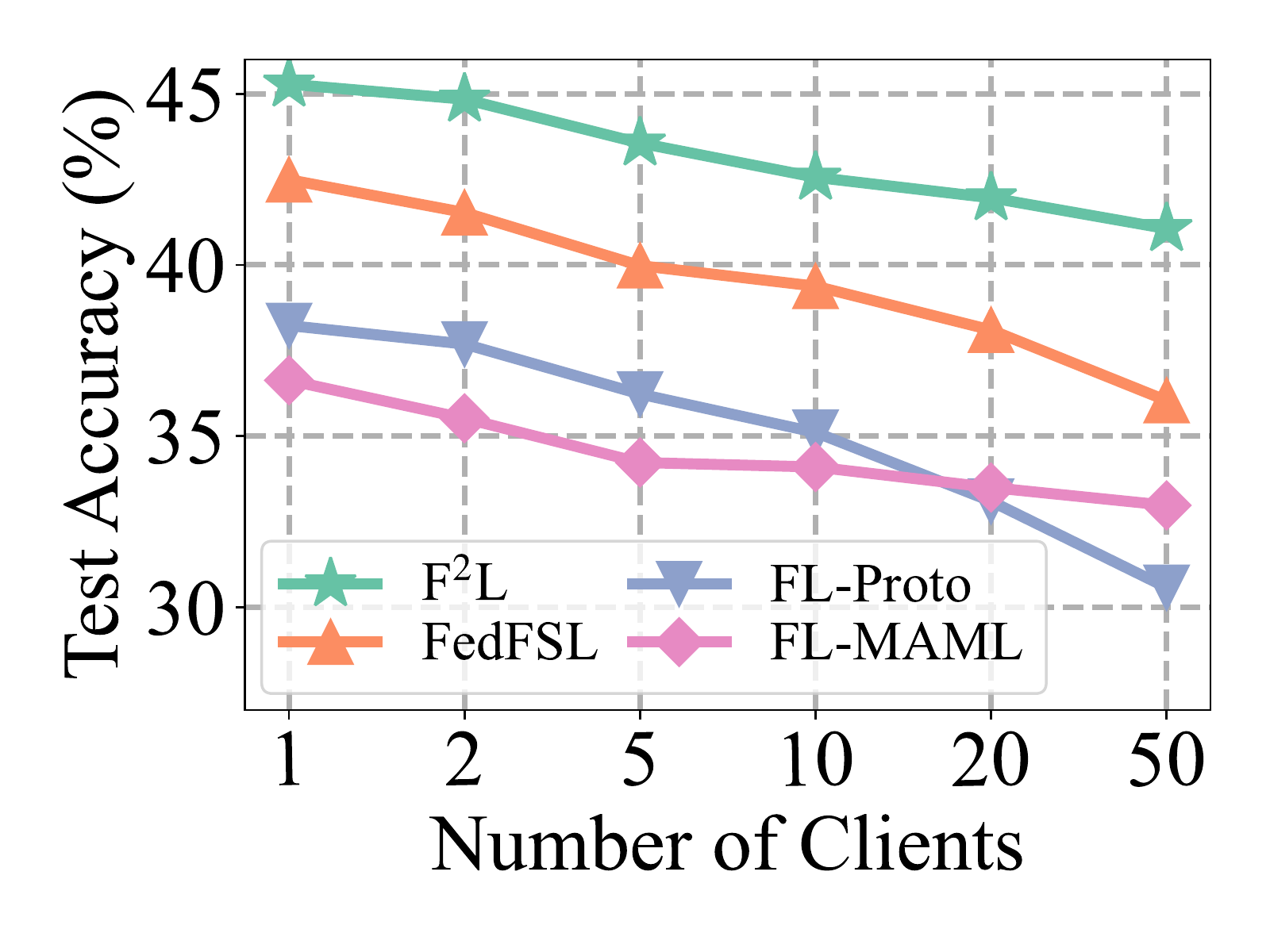}}
\subcaptionbox{5-shot}
{\includegraphics[width=0.23\textwidth]{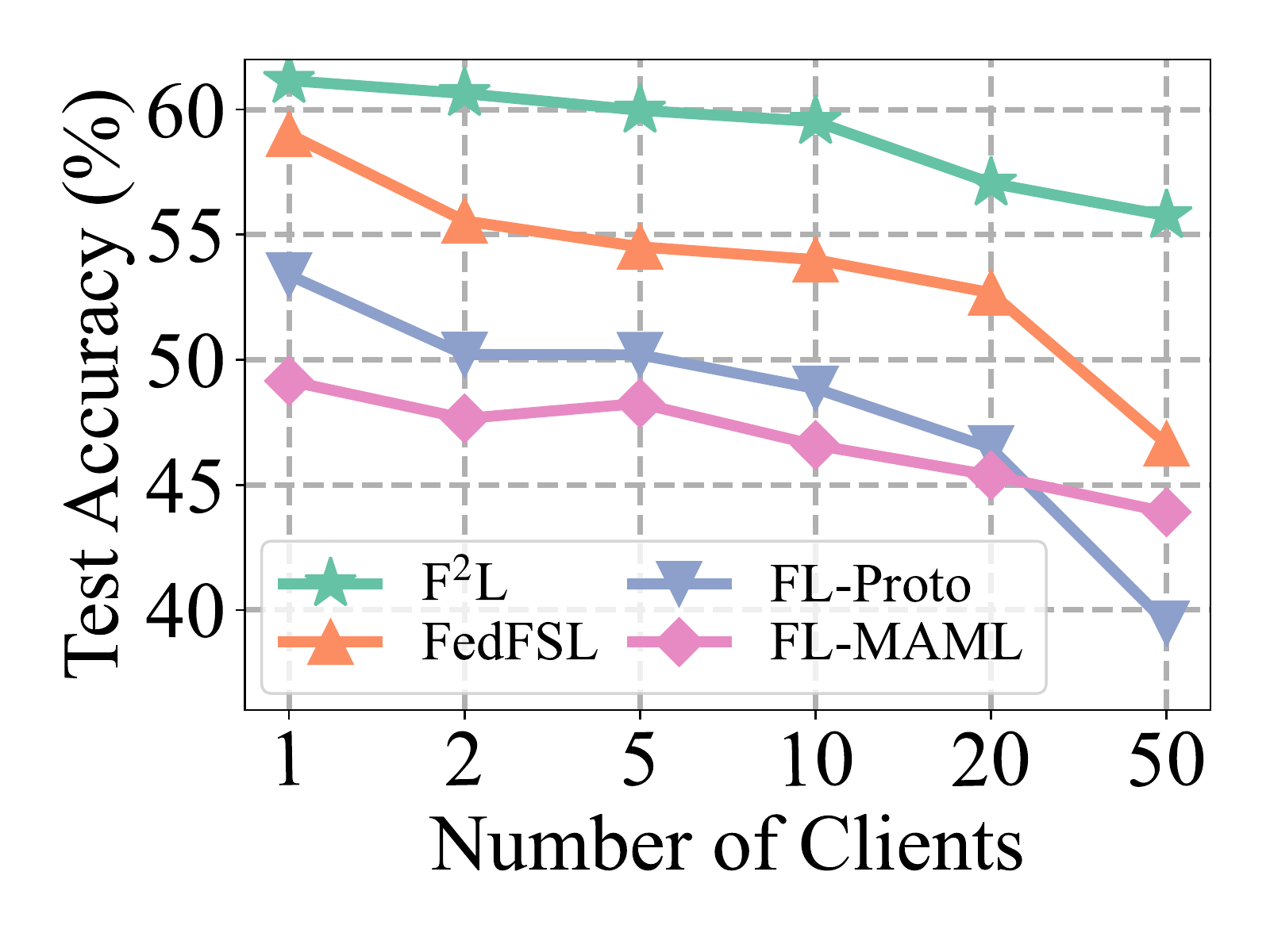}}
\vspace{-0.1in}
		\caption{The results of non-IID federated 1-shot and 5-shot learning on FC100 regarding the number of clients.}
		\label{fig:client_number}
\vspace{-0.2in}
	\end{figure}
	
\subsubsection{Effect of Client Number}
In this section, we study the robustness of our framework under the scenario with a varying number of clients. In particular, we keep the total training data unchanged, which means with more clients participating in the training process, each client preserves fewer training samples. As a result, the training performance will be inevitably reduced. 
Specifically, we partition the total training data into $I=1,2,5,10,20,$ and $50$ clients. Note that $I=1$ denotes the setting of completely centralized training. The results on FC100 with 1-shot and 5-shot settings are presented in Fig~\ref{fig:client_number} (we have similar results for other datasets and omit them for brevity). From the results, we can observe that all methods encounter a performance drop in the presence of more clients. Nevertheless, our framework F$^2$L can reduce the adverse impact brought by more clients through effectively leveraging the global knowledge learned from all clients. In consequence, the performance degradation is less significant for F$^2$L.

\section{Related Work}	
\subsection{Few-shot Learning}
The objective of Few-shot Learning (FSL) is to learn transferable meta-knowledge from tasks with abundant information and generalize such knowledge to novel tasks that consist of only limited labeled samples~\cite{sun2019meta,du2020few,wang2021reform,tan2022transductive,ding2020graph}. Existing few-shot learning works can be divided into two categories: \emph{metric-based} methods and \emph{optimization-based} methods. The metric-based methods target at learning generalizable metric functions to classify query samples by matching them with support samples~\cite{liu2019learning,sung2018learning, wang2022task}. 
For instance, Prototypical Networks~\cite{snell2017prototypical} learn a prototype representation for each class and conduct predictions based on the Euclidean distances between query samples and the prototypes. 
Relation Networks~\cite{sung2018learning} learn relation scores for classification in a non-linear manner. 
On the other hand, optimization-based approaches generally optimize model parameters based on the gradients calculated from few-shot samples~\cite{mishra2018simple,ravi2016optimization,khodak2019adaptive,wang2022glitter}. As an example, MAML~\cite{finn2017model} proposes to optimize model parameters based on gradients on support samples to achieve fast generalization. In addition, LSTM-based meta-learner~\cite{ravi2016optimization} adjusts the step size to adaptively update parameters during meta-training.
\subsection{Federated Learning}
Federated Learning (FL) enables multiple clients to collaboratively train a model without exchanging the local data explicitly~\cite{yang2019federated,kairouz2021advances,t2020personalized,li2019fedmd,zhu2021data,fu2022federated}. 
As a classic example, FedAvg~\cite{mcmahan2017communication} performs stochastic gradient descent (SGD) on each client to update model parameters and send them to the server. The server averages the received model parameters to achieve a global model for the next round. FedProx~\cite{li2020federated} incorporates a proximal term into the local update of each client to reduce the distance between the global model and the local model. To deal with the non-IID problem in FL, recent works also focus on personalization in FL~\cite{fallah2020personalized,arivazhagan2019federated,tan2022towards,bui2019federated}. For instance, FedMeta~\cite{chen2018federated} incorporates MAML~\cite{finn2017model} into the local update process in each client for personalization. FedRep~\cite{collins2021exploiting} learns shared representations among clients. Moreover, 
FedFSL~\cite{fan2021federated} proposes to combine MAML and an adversarial learning strategy~\cite{goodfellow2014generative,saito2018maximum} to learn a consistent feature space. 

\section{Conclusion}
In this paper, we study the problem of federated few-shot learning, which 
aims at learning a federated model that can achieve satisfactory performance on new tasks with limited labeled samples. 
Nevertheless, it remains difficult to perform federated few-shot learning due to two challenges: global data variance and local data insufficiency.
To tackle these challenges, we propose a novel federated few-shot learning framework F$^2$L. In particular, we handle global data variance by decoupling the learning of local meta-knowledge. Then we leverage the global knowledge that is learned from all clients to tackle the local data insufficiency issue. We conduct extensive experiments on four prevalent few-shot learning datasets under the federated setting, covering both news articles and images. The experimental results further validate the superiority of our framework F$^2$L over other state-of-the-art baselines. 

  \section{Acknowledgements}
The work in this paper is supported by the National Science Foundation under grants (IIS-2006844, IIS-2144209, IIS-2223769, CNS-2154962, and BCS-2228534), the Commonwealth Cyber Initiative awards (VV-1Q23- 007 and HV-2Q23-003), the JP Morgan Chase Faculty Research Award, the Cisco Faculty Research Award, the Jefferson Lab subcontract 23-D0163, and the UVA 4-VA collaborative research grant.

		\bibliographystyle{ACM-Reference-Format}
		\bibliography{acmart}

	\clearpage
	\begin{appendices}

\appendix
\section{Notations}\label{app:Notations}
In this section, we provide details for the used notations in this paper and their corresponding descriptions.
\begin{table}[htbp]
\setlength\tabcolsep{5pt}
\caption{Notations used in this paper.} 
\label{tb:symbols}
\begin{tabular}{cc}
\hline
\textbf{Notations}       & \textbf{Definitions or Descriptions} \\
\hline
$\mathbb{S}$& the server\\
$I$& the number of clients\\
$\mathbb{C}^{(i)}$& the $i$-th client\\
$\mathcal{C}_b$, $\mathcal{C}_n$ & the base class set and the novel class set\\
$\mathcal{D}_b^{(i)}$, $\mathcal{D}_n^{(i)}$& the local base and novel datasets in $\mathbb{C}^{(i)}$\\
$\mathcal{T}$, $\mathcal{S}$, $\mathcal{Q}$&a meta-task and its support set and query set\\
$\psi$, $\phi$& the client-model and the server-model\\
$\bh_\psi$&representations learned by client-model\\
$\bh_\phi$&representations learned by server-model\\
$\bp_\psi$& the output probabilities by client-model\\
$\bp_\phi$& the output probabilities by server-model\\
$N$&the number of support classes in each meta-task\\
$K$&the number of labeled samples in each class\\
$D$& the number of support samples in each meta-task\\
$Q$& the number of query samples in each meta-task\\
$\alpha_{ft}$ &the learning rate for fine-tuning\\
$\alpha_\phi$, $\alpha_\psi$&the meta-learning rates for $\phi$ and $\psi$\\
$T$& the number of training rounds\\
$\tau$& the number of local training steps\\
$\lambda_{MI}$& the loss weight for the mutual information loss\\
$\lambda_{KD}$& the loss weight for the knowledge distillation loss\\
\hline
\end{tabular}
\end{table}

\section{Algorithm}
We provide the detailed training process of our framework F$^2$L in Algorithm~\ref{algorithm}.
	\begin{algorithm} [htbp]
		\caption {Detailed training process of our framework F$^2$L.}
		\begin{algorithmic}[1]
			\REQUIRE A set of $I$ federated clients; a local update objective $\mathcal{L}_{\phi}$ for server-model; a local update objective $\mathcal{L}_{\psi}$ for client-model; number of training rounds $T$; number of local training steps $\tau$.
			\ENSURE A trained server-model $\phi$ and a unique client-model $\psi_i$ for each client $\mathbb{C}^{(i)}$ in $\{\mathbb{C}^{(i)}\}_{i=1}^I$.
			\FOR {$t=1,2,\dotsc,T$}
			\FOR {each client $\mathbb{C}^{(i)}$ in $\{\mathbb{C}^{(i)}\}_{i=1}^I$ in parallel}
			    \FOR {$s=1,2,\dotsc,\tau$}
                \STATE Sample a meta-task $\mathcal{T}_{i}^{t,s}=\{\mathcal{S}_{i}^{t,s},\mathcal{Q}_{i}^{t,s}\}$;
                \STATE Fine-tune client-model $\psi_i$ on $\mathcal{S}_{i}^{t,s}$ according to Eq. (\ref{eq:ft});
                \STATE Update server-model on $\mathcal{S}_{i}^{t,s}$ with Eq. (\ref{eq:all_model}) and Eq. (\ref{eq:transfer_objective});
                \STATE Update client-model on $\mathcal{Q}_{i}^{t,s}$ with Eq. (\ref{eq:few_model}) and Eq. (\ref{eq:achieve_objective});
			    \ENDFOR
			\ENDFOR
			\STATE Each client returns the updated parameters of server-model to the server;
			\STATE The server sends back averaged parameters of server-model to each client;
			\ENDFOR
			
		\end{algorithmic}
					\label{algorithm}
	\end{algorithm}

\section{Reproducibility}
\subsection{Model Details}
\label{app:encoder}
In this section, we introduce the specific choices for the encoders and classifiers in both server-model and client-model (i.e., $q_\phi$, $f_\phi$, $q_\psi$, and $f_\psi$). 
\subsubsection{Server-model Encoder $q_\phi$}
For the server-model encoder, we adopt different models for news article datasets and image datasets. In particular, for news article datasets 20 Newsgroup and Huffpost, we leverage a biLSTM~\cite{hochreiter1997long} with 50 units as the server-model encoder. For the image datasets FC100 and miniImageNet, following~\cite{tian2020rethinking,ravichandran2019few}, we utilize a ResNet12 as the server-model encoder. Similar to~\cite{lee2019meta}, the Dropblock is used as a regularizer. The number of filters is set as (64, 160, 320, 640).
\subsubsection{Client-model Encoder $q_\psi$}
 Considering that the client-model is required to process the entire support set in a meta-task for 
 learning local meta-knowledge, we propose to further utilize a set-invariant function that takes a set of samples as input while capturing the correlations among these samples. In practice, we leverage the Transformer~\cite{vaswani2017attention} as the client-model encoder $q_\psi$ to process the entire support set:
\begin{equation}
\left(\mathbf{h}^1_\psi,\mathbf{h}^2_\psi,\dotsc,\mathbf{h}^{D}_\psi\right)=\text{Transformer}\left(\mathbf{h}^1_\phi,\mathbf{h}^2_\phi,\dotsc,\mathbf{h}^{D}_\phi\right),
\end{equation}
where $\bh^i_\phi$ (or $\bh^i_\psi$) denotes the representation of the $i$-th sample in $\mathcal{S}$ learned by the server-model encoder $q_\phi$ (or client-model encoder $q_\psi$). With the Transformer, the representations learned by the client-model can effectively capture the correlations among samples in the entire support set $\mathcal{S}$ for learning meta-knowledge.

\subsubsection{Server-model Classifier $f_\phi$ and Client-model Classifier $f_\psi$}
The classifiers $f_\phi$ and $f_\psi$ are both implemented as a fully-connected layer, where the output size is $|\mathcal{C}_b|$ for $f_\phi$ and $N$ for $f_\psi$, as described in Sec.~\ref{sec:overall}.

\subsection{Baseline Settings}
In this section, we provide further details in the implementation of baselines in our experiments.
	\begin{itemize}
    \item  \emph{Local}. For this baseline, an individual model is trained for each client over the local data. Specifically, we use the same architecture of encoders in our framework to learn sample representations.
    \item \emph{FL-MAML}. For this baseline, we leverage the MAML~\cite{finn2017model} strategy and set the meta-learning rate as 0.001 and the fine-tuning rate as 0.01. The encoders are the same as our framework.
    \item \emph{FL-Proto}. For this baseline, we follow the setting in ProtoNet~\cite{snell2017prototypical} with the same encoders in our framework. The learning rate is set as 0.001.
    \item \emph{FedFSL}~\cite{fan2021federated}. For this baseline, which combines MAML and an adversarial learning strategy~\cite{goodfellow2014generative,saito2018maximum}, we follow the settings in the public code and set the learning rate as 0.001. The adaptation step size is set as 0.01.
 
\end{itemize}
\subsection{Parameter Settings}\label{app:parameter}
For our framework F$^2$L, we set the number of clients as 10. The number of training steps $\tau$ in each client is set as 10, and the number of training rounds $T$ is set as 200. Moreover, the meta-learning rates $\alpha_\psi$ and $\alpha_\phi$ are both set as 0.001 with a dropout rate of 0.1. The fine-tuning learning rate $\alpha_{ft}$ is set as 0.01. We leverage the Adam~\cite{kingma2014adam} optimization strategy with the weight decay rate set as $10^{-4}$. During the meta-test, we randomly sample 100 meta-test tasks from novel classes $\mathcal{C}_n$ with a query set size $|\mathcal{Q}|$ of 5. In order to preserve consistency for fair comparisons, we keep identical meta-test tasks for all baselines. The loss weights $\lambda_{MI}$ and $\lambda_{KD}$ are both set as 0.5. The default value of $I$ is set as 10.

  \end{appendices}
\end{document}